\documentclass[runningheads]{llncs}
\usepackage{graphicx}
\usepackage{amsmath}
\usepackage{subfig}
\begin{document}
\title{One-Class Classification for Wafer Map using Adversarial Autoencoder with DSVDD Prior}
%
%
\author{Ha Young Jo
\and Seong-Whan Lee}

\authorrunning{H. Jo et al.}
%
\institute{Department of Artificial intelligence\\
Korea University, Seoul 02841, Republic of Korea\\
\email{\{hayoungjo,sw.lee\}@korea.ac.kr}}
%
\maketitle              
\begin{abstract}
Recently, semiconductors' demand has exploded in virtual reality, smartphones, wearable devices, the internet of things, robotics, and automobiles. Semiconductor manufacturers want to make semiconductors with high yields. To do this, manufacturers conduct many quality assurance activities. Wafer map pattern classification is a typical way of quality assurance. The defect pattern on the wafer map can tell us which process has a problem. Most of the existing wafer map classification methods are based on supervised methods. The supervised methods tend to have high performance, but they require extensive labor and expert knowledge to produce labeled datasets with a balanced distribution in mind. In the semiconductor manufacturing process, it is challenging to get defect data with balanced distribution. In this paper, we propose a one-class classification method using an Adversarial Autoencoder (AAE) with Deep Support Vector Data Description (DSVDD) prior, which generates random vectors within the hypersphere of DSVDD. We use the WM-811k dataset, which consists of a real-world wafer map. We compare the F1 score performance of our model with DSVDD and AAE.

\end{abstract}
\section{Introduction}
The semiconductor market has been growing rapidly with the fourth industrial revolution \cite{jelinek2018global,batra2018artificial,shin2017r}. Semiconductors have been required in many areas such as virtual reality, smartphones, wearable devices, the internet of things (IoT), robotics, and automobiles \cite{monch2018survey1,uzsoy2018survey2,monch2018survey3}. A lot of types of semiconductors are demanded at diverse products in many areas. According to this demand, semiconductor manufacturing lines have more diversity, and the manufacturing process becomes more complicated. Semiconductor manufacturers want to make semiconductors with high yields for reinforcing or, at least, maintaining market competitiveness. To do this, semiconductor manufacturers do quality assurance activities like facility diagnosis, process control, stabilization of yield rate, etc. \cite{hsieh2019building,kao2018impact,mohn2017system,shon2021unsupervised}.

After the fabrication process, wafers are tested to check their electrical characteristics at the die on the wafer \cite{mann2004leading}. After the test, the results are represented as an image with information about dies that are malfunctioned and where the defect dies are located. We call this image a \textit{wafer map}. The defects on the wafer map can be categorized into some classes with spatial defect similarity: Loc, Edge-Loc, Edge-Ring, Center, Scratch, Random, Near-full, Donut, None. Some defect classes can help engineers in the manufacturer to recognize specific problems in the process \cite{hansen1997monitoring,yuan2011detection}. For example, inappropriate wafer handling manipulation may arise to a scratch pattern, layer misalignment during the storage node process can cause edge-ring pattern, and a uniformity problem at the chemical-mechanical planarization may make center pattern \cite{yuan2011detection}.

The basic method for classification of \textit{wafer map} is statistical methods. This method uses a statistical model with features extracted manually. However, traditional statistical methods have a limitation on scalability and nonlinearity. Recently, the deep neural methods of diverse architecture are being successfully applied in various area \cite{yang2007reconstruction,roh2007accurate,bulthoff2003biologically,ahmad2006human,lee1997new}. In accordance with this trend, deep neural networks are applied in semiconductor manufacturing.

Traditionally, supervised methods are used for wafer map classification. These methods require a lot of labels made by visual check from skillful engineers with their experience, knowledge, and intuition\cite{shankar2005defect}. However, Annotating label to a large number of wafer maps manually is expensive and takes a lot of time\cite{baly2012wafer}. Furthermore, when the engineers work long hours, they may feel fatigued, and it can cause miss classification. \cite{tan2011automated}. Therefore there are a lot of requirements for automated wafer map pattern classification.

There are two types of problems for wafer map classification that are actively researched. First problem is to handle the imbalanced dataset. After the process has stabilized, defect wafers hardly occur. For this reason, it is hard to get defect data. Furthermore, it is almost impossible to get balanced defect data over defect type. To overcome this problem, one common way is to generate artificial data containing defective images using generative models before the training stage. For example, in \cite{ji2020using}, researchers used a deep convolutional generative model for generating faulty images. And, in \cite{liu2021focal}, researchers used auxiliary classifier generative adversarial network with focal loss at discriminator and generated images for learning.

Second problem is the difficulty of getting a labeled data. Many semiconductor manufacturers try to apply the latest technology like deep learning, artificial intelligence. However, data that manufacturers already have is not preferred to apply the technology. Moreover, most of these data are without the label; thus, it needs additional labor for assigning the label. There are several ways to get over this problem. The first one is using active learning. Active learning is the way to use small labeled data and a lot of unlabeled data. This model learns with small labeled data and selects the most ambiguous data in the unlabeled dataset. Engineers or something called Oracle are then asked to label the selected ambiguous data. Then repeat the process by adding data labeled by Oracle until certain conditions are satisfied. In \cite{koutroulis2020enhanced}, researchers select the most ambiguous data using uncertainty and Density-based Spatial Clustering of Application with Noise (DBSCAN). The other one is using clustering methods. 

In this paper, we proposed an adversarial autoencoder-based one-class classification method. One-class classification learns a normal data distribution at the training stage and determines whether the input is similar to trained data or not. To do this, we use an adversarial autoencoder with DSVDD prior, which generates a random vector within the hypersphere. We use the public dataset of wafer map, called WM-811K ~\cite{wu2014wafer}, for performance evaluations and compare our model with AAE and DSVDD with identical architecture.
\section{Related works}
\subsection{Wafer map classification}
Many researchers have studied automated wafer map classification. In many studies, they use the raw wafer map as an input to classify the wafer defect. However, using the raw wafer map directly as an input is inappropriate because the raw wafer map does not have an effective characteristic for classification. There are two approaches for extracting effective features: the manual feature extraction approach and the deep learning-based approach.

Manual feature extraction approaches change wafer map images to characteristic vectors to effectively represent the discriminative features. It is already used in a real-world manufacturer. In \cite{wu2014wafer}, researchers conducted classification using geometry and radon-base features with a support vector machine (SVM). In \cite{fan2016wafer}, researchers conducted multi-label classification using the ``ordering point to identify the cluster structure'' and density- and geometry-based features with SVM. In \cite{piao2018decision}, the authors used an ensemble technique with radon-based features. In \cite{saqlain2019voting}, researchers suggested an ensemble classifier with various features such as density-, geometry-, and radon-based features.

Due to the success of deep learning in many areas, deep learning methods have been applied to wafer map classification tasks. These approaches use the raw wafer map as an input and automatically extract features. In \cite{nakazawa2018wafer}, researchers used CNN for wafer classification first time and generated a synthetic wafer map using the Poisson point process. In \cite{nakazawa2019anomaly}, the authors used a convolutional autoencoder for detecting and segmenting defect patterns. \cite{yu2019stacked} used a stacked convolutional sparse denoising autoencoder for extracting robust features. In \cite{kong2020semi}, researchers used the Ladder Network and semi-supervised classification framework for using labeled and unlabeled datasets. In \cite{kim2021wafer}, the authors used out-of-distribution data, which is undefined in the training dataset, for improving classification performance.

\subsection{Deep Support Vector Data Description}
One-class support vector machine (OC-SVM) is the classic one-class classification method. It is a special type of SVM that holds labels at the training stage.
Support Vector Data Description (SVDD) \cite{tax2004support} is advanced version of one-class SVM. SVDD maps all the training data consisting of only normal into kernel space and finds the hypersphere containing all the data points in the kernel space with the smallest volume. After the training stage, the model finds the hypersphere's center and radius, satisfying the condition, and we can detect anomalies when the data is outside of the hypersphere.

Deep SVDD \cite{ruff2018deep} is inspired SVDD and changes the kernel in SVDD to a deep neural network. The neural network is trained to have the smallest hypersphere that contains the latent vector of data in the latent space, with loss, Equation~\ref{eq: DSVDD loss}.
\begin{equation}
\label{eq: DSVDD loss}
    \min_{R,\mathcal{W}} R^{2} +\frac{1}{\nu n}\sum_{i=1}^{n}\max\{0,||\phi(\mathbf{x}_i;\mathcal{W})-\mathbf{c}||^{2}-R^{2}\}+\frac{\lambda}{2}\sum_{l=1}^{L}||\mathbf{W}^{l}||^{2}_{F}
\end{equation}
where, $R$ is the radius of the hypersphere, $\mathbf{c}$ is the center of the hypersphere, $\mathbf{W}^{l}$ is the weights at the layer $l$, $\mathcal{W}$ is the set of weights, $\mathcal{W}=\{\mathbf{W}^{1},\cdots,\mathbf{W}^{L}\}$, $\mathbf{x}_{i}$ is the $i$th input data, and $\nu, \,\lambda$ are hyperparameters. The second term is zero when the latent vector is within the hypersphere; therefore, it forces to be inside the hypersphere.

\subsection{Generative adversarial network}
Generative adversarial network (GAN) is the generative model that uses generator and discriminator. Discriminator determines whether the input is the real data or generated from the generator. The generator is encouraged to fool the discriminator by generating fake data that is as close to the real data as possible. At the training stage, The network is optimized by Equation~\ref{eq: GAN loss}.
\begin{equation}
\label{eq: GAN loss}
\max_{D} \min_{G}(\theta_{G},\theta_{D})=E_{x \sim P_{data}(x)}[log D(X)]+E_{x \sim P_{z}(X)}[1-log(D(G(z)))]
\end{equation}
where $D, G$ represent discriminator and generator, respectively, $\theta$ is the parameter of generator or discriminator network, $z$ is the random noise vector that is used as an input to the generator. We use the Adversarial autoencoder (AAE) for experiments. AAE is a network that combines autoencoder with adversarial learning. A discriminator in AAE determines whether the latent vector is from the real data or not, making the encoder acts as a generator in the GAN. The difference between GAN and AAE loss is adding an L1 loss for reconstruction. The total loss of AAE is the Equation~\ref{eq: AAE loss G}

\begin{equation}\label{eq: AAE loss G}
\begin{split}
\max_{D} \min_{e,d}(\theta_{e},\theta_{d},\theta_{D})= & E_{x \sim P_{data}(x)}[log D(e(X))]+E_{x \sim P_{z}(X)}[1-log(D(z))]\\
& +L1(X,d(e(x)))
\end{split}
\end{equation}
where $e, d$ is encoder and decoder, $L1$ is L1 loss.

\newpage
\section{Methods}
\subsection{DSVDD prior}\label{sect: DSVDD prior}
We want the distribution of the latent vector of the data to be within the DSVDD hypersphere. A vector extracted from the prior must be within the hypersphere or close to the boundary of the distribution to satisfy such a condition. Such a prior is then a DSVDD prior.
DSVDD prior is considered DSVDD distribution as prior. DSVDD distribution is the distribution that sampled random vector is about hypersphere of DSVDD. Latent vectors of data are forced by DSVDD to be concentrated at the center of the hypersphere of DSVDD. The hypersphere is a sphere without distortion. Therefore we can assume that DSVDD distribution is isotropic Gaussian distribution with the mean at the center of the hypersphere. We can formulate the random vector as Equation~\ref{eq:latent vector distribution}.
\begin{equation}
\label{eq:latent vector distribution}
z_i \sim \mathcal{N}(c_{i},\,\sigma^{2})=c_{i}+\mathcal{N}(0,\,\sigma^{2})
\end{equation}
where $z_i$ is $i$th latent vector component, $c_i$ is the $i$th center vector of the hypersphere, and $\sigma$ is the standard deviation for Gaussian distribution. 

When a random vector is extracted from Equation~\ref{eq:latent vector distribution}, it cannot guarantee the distance between vector and the center of the hypersphere is less than the radius of the hypersphere, or at least close to the radius. To overcome this problem, we need to handle the standard deviation in the Gaussian distribution. The expectation of the square distance can be calculated using Equation~\ref{eq: expectation distance}
\begin{equation}
\label{eq: expectation distance}
E(Z^{2})=\sum_{i=1}^{N}E(z_{i}^{2})=\sum_{i=1}^{N}(\mu^{2}+\sigma^{2})=N\sigma^{2}
\end{equation}
where $Z$ is the random vector, $N$ is the dimension of the latent space, $\mu, \sigma$ are mean and variance of the Gaussian distribution. We use the fact that mean is 0 at final term.\\
We want the expected distance to be the radius of the hypersphere. Like Equation~\ref{eq: sigma}, the variance become the value $R^2/N$

\begin{equation}\label{eq: sigma}
\begin{split}
E(\mathbf{Z}^{2}) & =N\sigma^{2}=R^{2}\\
\sigma^{2} & =\frac{R^{2}}{N}                        
\end{split}
\end{equation}
where $R$ is the radius of hypersphere.

Than total distribution of DSVDD is Equation ~\ref{eq: DSVDD distribution}
\begin{equation}
    \label{eq: DSVDD distribution}
    \mathbf{Z} \sim \mathbf{C}+\mathcal{N}(0,R^{2}/N\nu^2)
\end{equation}
where $\mathbf{Z}$ is the random vector, $\mathbf{C}$ is the center of the hypersphere, and $\nu$ is hyper parameter for handling the expectation distance.

\subsection{Assign label of random vector}
In ordinary generative models like GANs, the random vectors are labeled as being normal when the discriminator is learned since every random vector is generated by prior. In our model's prior, we expect all random vectors are within the hypersphere. However, in actuality, all random vectors extracted from Equation~\ref{eq: DSVDD distribution} are not always in the hypersphere. Therefore we need to assign the label to the random vectors. 

To assign the label, we use the DSVDD classification methods. If the random vector is out of the hypersphere, then the \textit{Real} label is assigned during the learning stage of the discriminator. If the random vector is in the hypersphere, then the \textit{Fake} label is assigned.
\begin{equation}\label{eq: assign label}
    \begin{aligned}
        l(\mathbf{Z},\mathbf{C})=\Big\{
        \begin{matrix}
                \textit{Fake} \quad \textrm{if}\quad \textrm{dist}(\mathbf{Z},\mathbf{C}) \le R\\
                \textit{Real} \quad \textrm{if}\quad \textrm{dist}(\mathbf{Z},\mathbf{C}) > R
        \end{matrix}
    \end{aligned}
\end{equation}  

\subsection{Architecture}
The network architecture used in experiments is based on adversarial autoencoder. The network has encoder and decoder like variational autoencoder, and discriminator for adversarial learning. 

All parts of the network are consist of linear layers. The encoder has separated two linear layers at the final layer to generate mean and variance vectors used for variational inference. The input images are consist of 3 channels; each channel represents margin, normal, and defect. At the pixel in the input image, only one of the channels has value 1 and others have 0. Due to the characteristics of input images, we added a softmax layer to the last layer of the decoder. And other components of the network are identical to the adversarial autoencoder.
The model needs to map the training data inside the hypersphere. To do this, we use AAE loss, Equation~\ref{eq: AAE loss G} with a DSVDD loss, Equation~\ref{eq: DSVDD loss}. On the other hand, we use the loss as an anomaly score for classification.

\section{Experiments}
\subsection{Data Description}
\begin{table}[]
\begin{center}
\caption{Description of WM-811K dataset}\label{table:data_description}
\begin{tabular}{|c|c|r|}
\hline
\textbf{Unlabeled} & \multicolumn{2}{r|}{\textbf{638,570}} \\ \hline
\textbf{Labeled}   & \multicolumn{2}{r|}{\textbf{172,950}} \\ \hline
                   & None               & 147,431          \\ \cline{2-3} 
                   & Loc                & 3,593            \\ \cline{2-3} 
                   & Edge-Loc           & 5,1589           \\ \cline{2-3} 
                   & Edge-Ring          & 9,680            \\ \cline{2-3} 
                   & Center             & 4,294            \\ \cline{2-3} 
                   & Scratch            & 1,193            \\ \cline{2-3} 
                   & Random             & 866              \\ \cline{2-3} 
                   & Near-full          & 149              \\ \cline{2-3} 
                   & Donut              & 555              \\ \hline
\textbf{Total}     & \multicolumn{2}{r|}{\textbf{811,520}} \\ \hline
\end{tabular}
\end{center}
\end{table}
For experiments, we used WM-811K dataset, which is real-world trace data obtained from a semiconductor manufacturer ~\cite{wu2014wafer}. It contains about 800,000 images with both labeled and unlabeled data. Around 630,000 of the data is without the label, and 170,000 of the data is with the label. There are nine distinctive labels with uneven distribution. You can see the details in Table ~\ref{table:data_description}.

\begin{figure}
\centering
  \subfloat[Loc]{%
    \includegraphics[width=.33\textwidth]{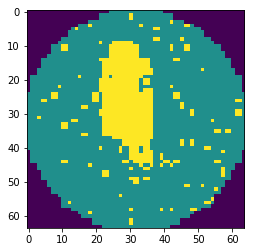}}\hfill
  \subfloat[Edge Loc]{%
    \includegraphics[width=.33\textwidth]{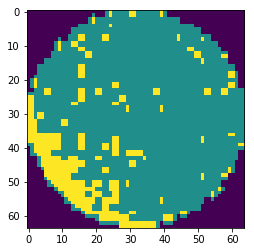}}\hfill
  \subfloat[Edge Ring]{%
    \includegraphics[width=.33\textwidth]{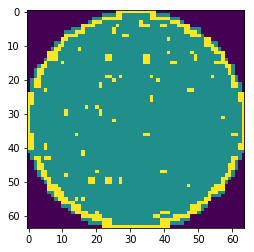}}\\
  \subfloat[Center]{%
    \includegraphics[width=.33\textwidth]{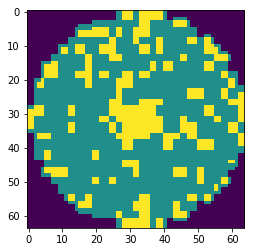}}\hfill
  \subfloat[Scratch]{%
    \includegraphics[width=.33\textwidth]{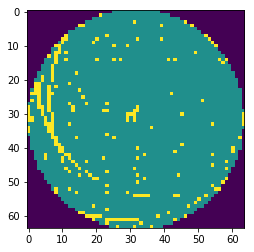}}\hfill
  \subfloat[Random]{%
    \includegraphics[width=.33\textwidth]{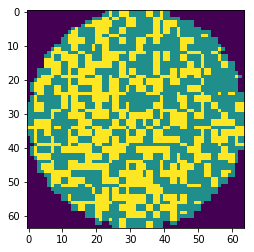}}\\
  \subfloat[Near Full]{%
    \includegraphics[width=.33\textwidth]{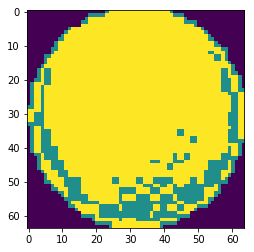}}\hfill
  \subfloat[Donut]{%
    \includegraphics[width=.33\textwidth]{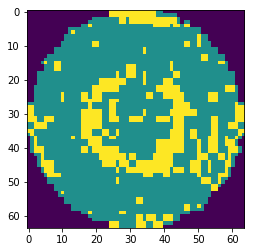}}\hfill
  \subfloat[None]{%
    \includegraphics[width=.33\textwidth]{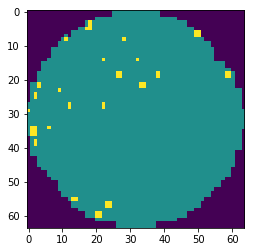}}\\
  \caption{Example of WM-811k dataset}\label{fig: wafer example}
\end{figure}

Engineers can infer the root cause of the device by defect pattern class. In this class, \textit{None} pattern is a normal pattern, and others are defect pattern. You can see the example of label of Wafer Bin Map of each type in Fig ~\ref{fig: wafer example}.

In this experiment, we used only normal data at the training stage, like other one-class classification frameworks. Then the data with defects are used at the validation and test stage. We re-labeled all defects patterns into one class since our task is only to classify normal data and defect data.\\
We split the data into training, validation, and test data. There is only \textit{none} type data in the training set and 80\% of all \textit{none} type data. There is 10\% of all \textit{none} type data in the validation set, and there are other patterns with 50\% of all each pattern. The rest of the datasets were used as test sets. You can show the detail in the Table ~\ref{table:experiment data}.

\begin{table}[]
\begin{center}
\caption{Number of experiment data.}\label{table:experiment data}
\begin{tabular}{|c|c|c|c|}
\hline
\textbf{Failure Type} & \textbf{Train} & \textbf{Validation} & \textbf{Test} \\ \hline
None                  & 117,944     & 14,743            & 14,744          \\ \hline
Loc                   & 0           & 1,796             & 1,797          \\ \hline
Edge-Loc              & 0           & 2,594             & 2,595          \\ \hline
Edge-Ring             & 0           & 4,840             & 4,840          \\ \hline
Center                & 0           & 2,147             & 2,147          \\ \hline
Scratch               & 0           & 596               & 597          \\ \hline
Random                & 0           & 433               & 433          \\ \hline
Near-full             & 0           & 74                & 75          \\ \hline
Donut                 & 0           & 277               & 278          \\ \hline
\textbf{Total}        & \textbf{117,944}  & \textbf{27500}       & \textbf{27506} \\ \hline
\end{tabular}
\end{center}
\end{table}

\subsection{Data processing}
The WM-811K dataset we used in this experiment has wafer images that represent whether dies on the wafer are defective or not. Each pixel on the images has only three types of value. If the pixel has 0 value, the pixel is out of the wafer. If the pixel has one value, the pixel represents a normal die. If the pixel has 2, the pixel represents defect die. These values that pixels have are categorical variables. Therefore we change the number of channels in images to 3, each channel represents out of the wafer, normal die, and defect die, like one-hot encoding.\\
We also need to match the size of the images since the images in the dataset have various sizes. We change the image size to 64 by 64.

\subsection{Comparing methods}
We selected DSVDD and AAE for comparing our model. Since our goal is to study the performance comparison between our loss function and other losses with the same architecture, we tried to maintain the model architecture as much as we can. The baseline architecture is selected based on the best-performing architecture of the AAE model. In the DSVDD, we used the encoder in the AAE model that outputs only the mean vector. In the AAE, we used the same architecture with our model without the SVDD loss and without assigning a label to the random vector.\\
To classify the data, whether it is normal or not, we used the loss of each model as an anomaly score. When the loss of DSVDD is more than 0, we assign the label \textit{defect}. Unlike the DSVDD case, the AAE model and our model are challenging to find an appropriate threshold. So we find the threshold as follows. First, we calculate the mean and standard deviation of the training set. 
We assumed that the threshold is close to the mean. After that, we find the upper limit of the anomaly score when the recall is 1 by increasing anomaly score by 0.1 time the standard deviation, from the mean. Then, we find appropriate threshold until the F1 score has highest value, by decreasing the anomaly score by 0.1 time the standard deviation, from the upper limit.

We used the threshold that have best performance at the validation dataset, to evaluate the test dataset. As an evaluation metric, we use accuracy, precision, recall, and F1 score. We input the label and inference result to functions of `metric' module in `sklearn' python package, accuracy\_score, precision\_score, recall\_score, f1\_score.
\section{Results}
Table~\ref{table: result} shows the results of the performance on the validation and test datasets. There is no significant difference between validation and test datasets. It might be the fact that the two datasets have the same ratio of the classes. Looking at each model, DSVDD showed the lowest performance except recall, accuracy, and precision are low under 0.5. It means that DSVDD can't map many normal data within the hypersphere. AAE has a similar recall with DSVDD but has high accuracy and precision score than DSVDD. This suggests that AAE maps normal data well to the center of the posterior rather than DSVDD mapping data to the center of the hypersphere. Our model has highest F1 score rather than others, but recall decrease. It might be the evidence that the DSVDD prior acts as constraint to get more \textit{none} data.
\begin{table}[]
\begin{center}
\caption{Results of experiments}\label{table: result}
\begin{tabular}{cc|c|c|c|c|c}
                                                     & \textbf{}           & \textbf{DSVDD} & \textbf{AAE} & \textbf{\begin{tabular}[c]{@{}c@{}}AAE\\ +DSVDD\\ /$\nu=1$\end{tabular}} & \textbf{\begin{tabular}[c]{@{}c@{}}AAE\\ +DSVDD\\ /$\nu=0.8$\end{tabular}} & \textbf{\begin{tabular}[c]{@{}c@{}}AAE\\ +DSVDD\\ /$\nu=1.2$\end{tabular}} \\ \hline\hline
                                                     & \textbf{Threshold} & 0              & 0.049        & 0.017                                                              & -0.674                                                               & 0.165                                                                \\ \hline\hline
\multicolumn{1}{c|}{{\textbf{Valid}}} & \textbf{Accuracy}    & 0.481855       & 0.640582     & 0.683564                                                           & 0.469127                                                             & 0.673345                                                             \\
\multicolumn{1}{c|}{}                                & \textbf{Precision}  & 0.468658       & 0.575095     & 0.621808                                                           & 0.466259                                                             & 0.623430                                                             \\
\multicolumn{1}{c|}{}                                & \textbf{Recall}     & 0.874422       & 0.86235      & 0.811319                                                           & 0.997648                                                             & 0.747119                                                             \\
\multicolumn{1}{c|}{}                                & \textbf{F1}         & 0.610246       & 0.690021     & 0.704034                                                           & 0.635508                                                             & 0.679693                                                             \\ \hline\hline
\multicolumn{1}{c|}{{\textbf{Test}}}  & \textbf{Accuracy}   & 0.477678       & 0.635716     & 0.679693                                                           & 0.469243                                                             & 0.679125                                                             \\
\multicolumn{1}{c|}{}                                & \textbf{Precision}  & 0.466719       & 0.570801     & 0.613425                                                           & 0.466374                                                             & 0.631060                                                             \\
\multicolumn{1}{c|}{}                                & \textbf{Recall}     & 0.881837       & 0.866087     & 0.822755                                                           & 0.998198                                                             & 0.742517                                                             \\
\multicolumn{1}{c|}{}                                & \textbf{F1}         & 0.610386       & 0.688103     & 0.702835                                                           & 0.635726                                                             & 0.682267                                                            
\end{tabular}
\end{center}
\end{table}

As we showed in Section~\ref{sect: DSVDD prior}, we will compare the results of three cases. The first case is when the random vector is generated from Equation \ref{eq: DSVDD distribution} with $\nu=0.8$. In this case, the most random vectors should be inside the hypersphere in DSVDD. The second case is when $\nu=1$ with which the random vector is generated near the hypersphere boundary. The last case is when $\nu=1.2$ with which the random vector is generated outside the hypersphere. Throughout the experiments, the values 0.8, 1.2 are empirically chosen. These values are the smallest values to generate the random vector inside and outside of the hypersphere.\\
In the first case, recall becomes nearly 1, and other metrics drop under 0.5. It means that almost all \textit{none} data is widely spread out from the center. The third case maintains the metrics except for recall. It means that more data is classified as normal because of the boundary expansion.

\section{Conclusion}
In this paper, we represented the way to integrate DSVDD and AAE by generating a random vector close to the hypersphere of DSVDD. Due to the random vector can be generated out of hypersphere, we assigned the label of the random vector when the discriminator is learned. And we confirmed that the F1 score has the highest value when some random vectors are generated in the hypersphere and other vectors are out of the hypersphere, comparing the case that all random vectors are in or out of the hypersphere.\\
We show that the DSVDD prior has more F1 score performance than AAE with the Gaussian prior when the network has the same architecture, but recall decrease. This fact tells us that the DSVDD prior acts as a constraint to get more \textit{none} data.\\
The future works of this paper are as follows. First, it needs to compare other one-class classification models. In this paper, we use AAE for one class classification based on a generative model, and DSVDD which is well known for anomaly detection. However, one-class classification is the research area studied in various fields, and there are many other recent models for one-class classification. Also, one class classification model based on a generative adversarial network is actively researched. Therefore, we need to check the performance of other models and chance to apply the idea to other adversarial models.\\
Second, we need to check the performance of each defect pattern. We want this model to apply real world, but it has low performance at the whole criterion. To improve the model's performance, we need to analyze miss classification. For the same reason, we should check the change of performance as a varying threshold. Threshold has an essential role in anomaly detection, and it is crucial to characterize performance as the threshold changes.

%
%
%
\bibliographystyle{splncs04unsrt}
\bibliography{reference.bib}

\end{document}